\documentclass{article}
\pdfoutput=1

\usepackage{arxiv}

\usepackage[utf8]{inputenc} 
\usepackage[T1]{fontenc}    
\usepackage{hyperref}       
\usepackage{url}            
\usepackage{booktabs}       
\usepackage{amsfonts}       
\usepackage{nicefrac}       
\usepackage{microtype}      
\usepackage{lipsum}		
\usepackage{graphicx}
\usepackage{natbib}
\usepackage{doi}

\usepackage{subfig}
\usepackage{multirow}
\usepackage{bbding}
\usepackage{floatrow}
\usepackage{amsmath}

\title{Cost Volume Pyramid Network with Multi-strategies Range Searching for Multi-view Stereo}

\author{ Shiyu Gao \\ Institute of Computing Technology, \\Chinese Academy of Sciences\\ gaoshiyu@ict.ac.cn \and
Zhaoxin Li  \\ Institute of Computing Technology, \\Chinese Academy of Sciences \and
Zhaoqi Wang \\Institute of Computing Technology, \\Chinese Academy of Sciences 
}

\renewcommand{\shorttitle}{\textit{arXiv} Template}

\begin{document}
\maketitle

\begin{abstract}
Multi-view stereo is an important research task in computer vision while still keeping challenging.
In recent years, deep learning-based methods have shown superior performance on this task.
Cost volume pyramid network–based methods which progressively refine depth map in coarse-to-fine manner, have yielded promising results while consuming less memory.
However, these methods fail to take fully consideration of the characteristics of the cost volumes in each stage, leading to adopt similar range search strategies for each cost volume stage.
In this work, we present a novel cost volume pyramid based network with different searching strategies for multi-view stereo.
By choosing different depth range sampling strategies and applying adaptive unimodal filtering, we are able to obtain more accurate depth estimation in low resolution stages and iteratively upsample depth map to arbitrary resolution.
We conducted extensive experiments on both DTU and BlendedMVS datasets, and results show that our method outperforms most state-of-the-art methods.
Code is available at: 
\url{https://github.com/SibylGao/MSCVP-MVSNet.git}
\end{abstract}

\keywords{Multi-view stereo  \and 3D reconstruction \and Cost volume \and Coarse-to-fine}

\section{Introduction}
Multi-view stereo is one of the fundamental computer vision tasks which is widely used in augmented reality, 3D modeling and autonomous driving. 
In deep learning era, deep CNNs used for cost regularization and extracting representative image features have achieved promising performance. Yao et al. \cite{yao2018mvsnet} first proposed an end-to-end MVS pipeline that constructs cost volume based on plane sweeping algorithm and aggregates different views by minimizing differential variance. However, this method consumes huge memory because that 3D CNN used for regularization is cubically proportional to image resolution. As a result, subsequent methods like \cite{yao2018mvsnet,yao2019recurrent} downsample high resolution images to regularize cost volume in a smaller resolution. To this end, methods designed in coarse-to-fine manner \cite{chen2019point,yu2020fast,yang2020cost,gu2020cascade} are put forward, which iteratively refine depth map based on cost volume pyramid and consume less memory. 

However, current coarse-to-fine methods suffer from two limitations.
First, the accuracy of the predicted depth map is highly dependent on the initial low-resolution depth map, since it is difficult to correct the depth of ill-posed and occluded pixels in the following narrow range. 
Second, current coarse-to-fine methods use same searching strategies in refinement stages after gaining initial depth map, which, however, not fully considered the characteristics of the cost volumes in each stage.

In this work, we propose a multi-strategies cost volume pyramid multi-view stereo network (MSCVP-MVSNet).
Instead of single depth range searching strategy, we utilize multi-dimensional information to calculate depth searching range for each layer.
To further utilize the information contained in the cost volume, we introduce unimodal distribution as a training label at second stage during the training process.

Our main contributions can be summarized as follows:

We present multiple depth range searching methods in different stages of pyramid structure, leveraging multi-dimension information.
On the second stage, variance-based strategy is applied to exploit previous predicted probabilities for each pixel.
For the succeeding refinement stage, we employ parameter-free method to propagate neighboring information to an arbitrary resolution during upsampling.

To further exploit information in cost volume of deep MVS and obtain more accurate predictions in low-resolution stage before refinement, we propose unimodal assumption as a training label in second stage.

Quantitative results show that our method obtains SOTA results on DTU dataset and satisfactory qualitative results on BlendedMVS.\\
\begin{figure}
	\begin{center}
		\includegraphics[width=0.6\columnwidth]{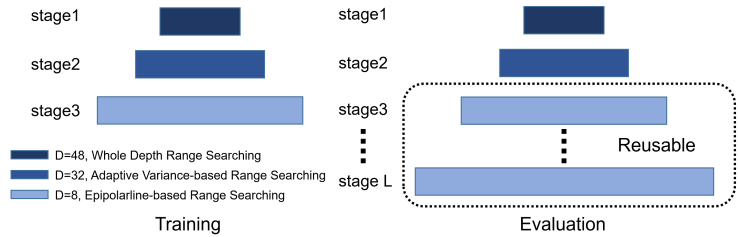}
	\end{center}
	\caption{Our method during training and evaluation.}
	\label{figure:MS}
\end{figure}
\section{Related Work}
\label{sec:2}
\subsection{Coarse-to-fine MVS methods.}
Deep MVS methods \cite{yao2018mvsnet,yao2019recurrent} based on pipeline of MVSNet \cite{yao2018mvsnet} build cost volume at the resolution of output images, which usually occupy large memory dealing with high resolution dataset such as DTU \cite{aanaes2016large} or Tanks and Temples \cite{knapitsch2017tanks}.
In order to save memory and computation consumption, coarse-to-fine methods \cite{yu2020fast,yang2020cost,gu2020cascade}are put forward.
For example, CVP-MVSNet \cite{yang2020cost} and Cascade-MVSNet \cite{gu2020cascade} build cost volume across the entire depth range in the coarsest resolution, after that, a narrowed sampling range is calculated based on previous depth predictions.
Based on these works \cite{yang2020cost,gu2020cascade}, Yu et al \cite{yu2021attention} propose AACVP-MVSNet, which introduces attention mechanism to CVP-MVSNet \cite{yang2020cost} framework.
Zhang et al. \cite{zhang2020visibility} took into account the visibility between different views based on Cascade-MVSNet \cite{gu2020cascade}.

\subsection{Depth sampling range.}
Coarse-to-fine pyramid networks uniformly sample the entire depth range in the first stage.
In the following stage, they iteratively narrow depth searching range by various strategies.
CVP-MVSNet \cite{yang2020cost} determines the local sampling range around the current depth by back projecting the corresponding pixels along epipolar line in source views.
Cas-MVSNet \cite{gu2020cascade} narrows sampling range of each stage by hand-crafted range with specific decay ratio.
For the first time, Cheng et al. \cite{cheng2020deep} utilized variance of probability distribution to describe the uncertainty of depth estimation.

All these methods mentioned above employ identical sampling range searching strategies in each stage of three- or four-layer pyramid.
In order to leverage both variance and neighbouring contextual information without adding complicated neural network modules, we apply different sampling range calculation strategies in different stage of coarse-to-fine MVS framework.

\subsection{Cost volume.}
Recently, cost volume is widely used in MVS and stereo matching methods.
MVSNet \cite{yao2018mvsnet} first introduces cost volume for end-to-end MVS pipeline by calculating photometric matching cost of each pixel in different fronto-parallel planes hypothesis.
A standard cost volume has a resolution of $H\times W\times D\times F$, where $H$, $W$, $D$, $F$ are height, width, number of plane hypothesis and feature channels, respectively.
While cost volume indicates matching cost of each depth hypothesis of each pixel intuitively, it is regularized by 3D UNet to generate an estimated probability value and indirectly supervised as an intermediate layer.
In order to integrate multi-scale information of cost volume, Shen et al. \cite{shen2021cfnet} proposed cost volume fusion module to obtain better initial disparity map.
Like CFNet \cite{shen2021cfnet}, we further utilize cost volume to obtain better initial depth map before refinement.
\begin{figure*}[!htb]
	\begin{center}
		\includegraphics[scale=0.14]{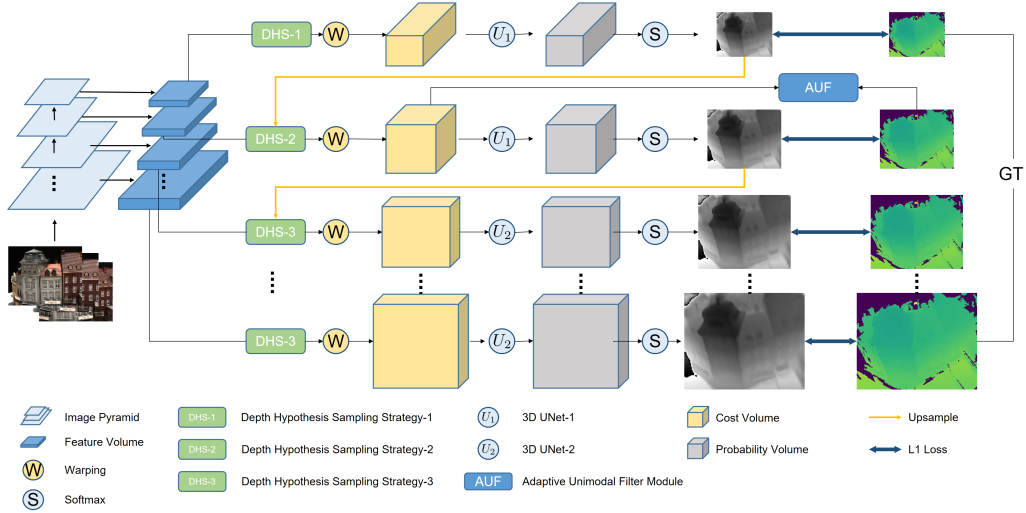}\\
		\caption{The network structure of MSCVP-MVSNet. }
		\label{figure:framework}
	\end{center}
\end{figure*}

\section{Methods}
\label{sec:3}
\subsection{Overview}
In this section,we introduce our multi-strategies cost volume pyramid network for high-resolution MVS reconstruction in details.
The overview of the network is shown in Fig. \ref{figure:framework}.
We assume the input reference image denoted by $I_{0}\in \mathbb{R}^{H\times W}$, and source images represented by $\{ I_{i}\}_{i=1}^{N-1}$.
To build a pyramidal structure, we downsample input images $L$ times to obtain images pyramid $\{I^{j}_{i}\}^{L}_{j=1}$, where $i \in \{0,1,\cdots,N\}$.
Feature pyramid $\{F^{j}_{i}\}^{L}_{j=1}$ are build by weights-shared feature extraction module.

As shown in Fig. \ref{figure:framework}, three different sampling strategies and two separated UNets are employed in our framework.

Inspired by GwcNet \cite{guo2019group}, we build cost volume by group-wise correlation instead of calculating feature volume variance over all views proposed by Yao et al. \cite{yao2018mvsnet}.

\begin{figure}[htb]
\centering
\subfloat[Before and after AUF in stage 2.]{\includegraphics[width=0.34\linewidth]{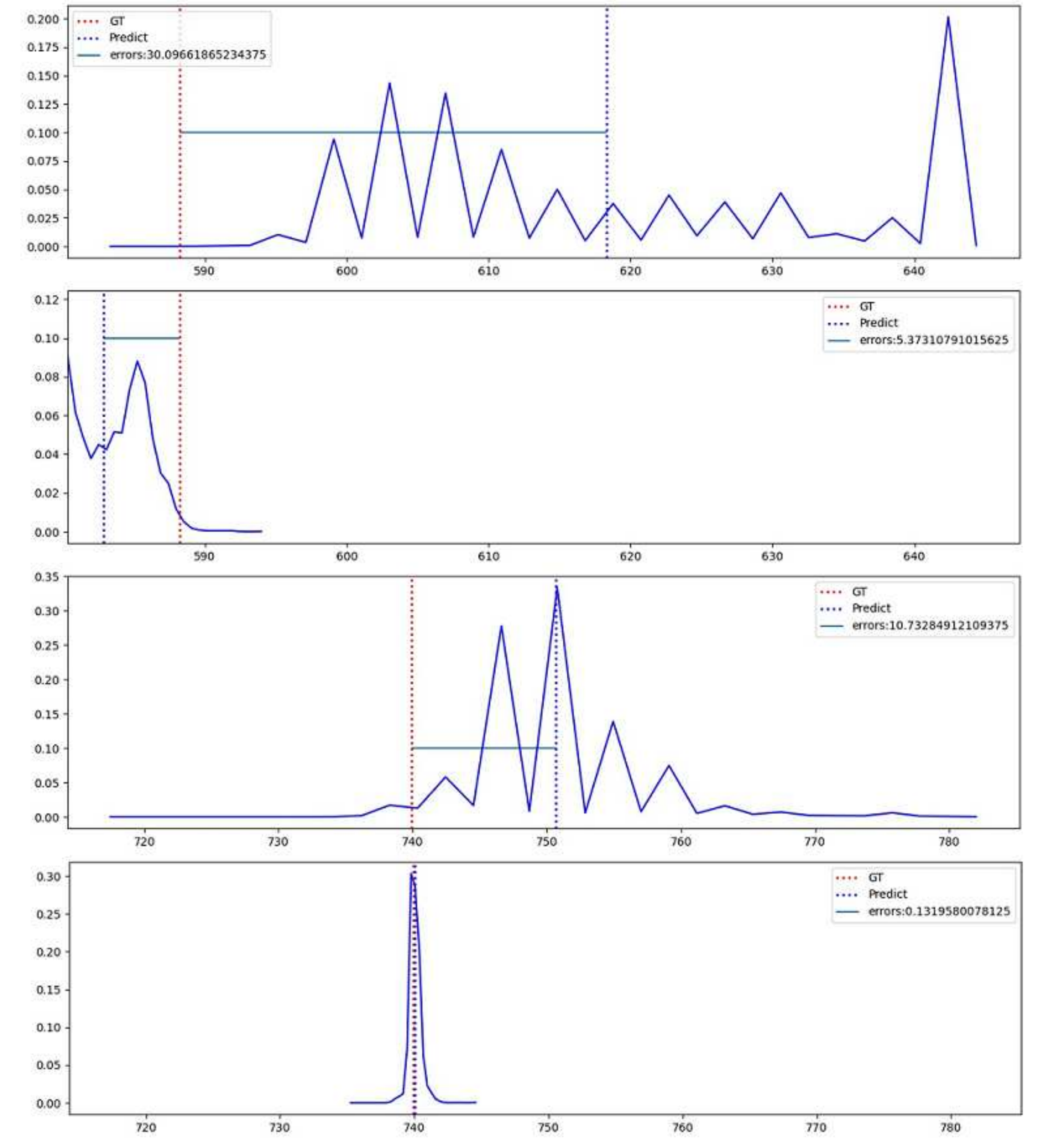}}
\hfill
\subfloat[Depth searching range in each stage]{\includegraphics[width=0.64\linewidth]{depth_range.pdf}}
\caption{Depth searching range visualization.}
\label{figure:multi-strategies}
\end{figure}

\subsection{Depth sampling range estimation}
As introduced in related work, previous methods (\cite{yang2020cost}, \cite{shen2021cfnet}, \cite{yu2021attention}) employ single strategy in each stage to calculate depth range, which either ignore statistical properties of each pixel or neighbouring information.
To solve this, we fuse multi-dimensional information by simply combine different uncertainty estimation strategies in different stage without any additional neural network modules and achieve satisfactory results.

In this section, we present our depth hypothesis sampling strategies in details.
As shown in Fig. \ref{figure:MS}, the number of pyramid layers in our framework is flexible, we train 3 different layers while evaluate with arbitrary number of stages.

In the first stage, we uniformly sampled depth hypothesis over the entire range to obtain a coarsest initial depth map.
Due to the large sampling range, we sampled more depth hypothesis ($D^1=48$) in this stage.

For second stage, we take advantage of probability distributions to calculate specific depth sampling range of each pixel.
Previous methods (\cite{zhang2020adaptive}, \cite{shen2021cfnet}) indicate that texture-less and occluded pixels tend to have multiple or wrong matches, as a result, the expectation of the per-pixel distributions can not depict the properties of multimodality and dispersion.
To solve this issue, we leverage the variance of the probability distribution as well as adaptive unimodal constraints (Sec. 3.3) to estimate per-pixel uncertainty and reduce local maxima of probabilities.
We set the number of depth hypothesis, $D^2=32$ in this stage.
For stage $l$, the variance at pixel $i$ is defined as:
\begin{equation}\label{eq1}
	\begin{aligned}
		\hat{V^{l}_{i}}=\sum_{j=1}^{D^{l}}P^{l}_{i,j}(d^{l}_{i,j}-\hat{d^{l}_{i}})^2,
	\end{aligned}
\end{equation}
where $P^{l}_{i,j}$ is the probability of pixel $i$ at sampled depth $j$,$d^{l}_{i,j}$ is the depth of sampled plane $j$, $\hat{d^{l}_{i}}$ is the estimated depth of pixel $i$ at current stage.

Different from UCSNet \cite{cheng2020deep}, we adopt the idea of CFNet[18] that originally proposed in stereo matching task, which use learned instead of hand-crafted scale parameters to determine confidence interval:
\begin{equation}%
	\begin{aligned}\label{eq2}
		d_{max}^{l+1}(i)=\hat{d^{l}_{i}} +\alpha^{l}\sqrt{ \hat{V^{l}_{i}} }+\beta^{l}, \\
		d_{min}^{l+1}(i)=\hat{d^{l}_{i}} -\alpha^{l}\sqrt{ \hat{V^{l}_{i}} }-\beta^{l},
	\end{aligned}
\end{equation}
where $\alpha^{l}$ and $\beta^{l}$ are learned parameters in stage $l$. Same as CFNet \cite{shen2021cfnet}, we initial $\alpha^l$ and $\beta^l$ as 0 at the beginning of training.
In texture-less regions with multi-modal distributions, the variances tend to be large, and adaptive uncertainty range estimation algorithm adjust depth hypothesis to a larger range so as not to miss the truth depth value before small-range refinement.
Depth searching ranges in Fig.\ref{figure:multi-strategies} show the effectiveness of our variance-based method in $2^{nd}$ stage.

Our first two layers have yielded fair results at the low resolution stage, and the depth values of high-resolution depth maps are obtained via upsampling operation.
Specifically, we apply parameters-free method to determine sampling range, which takes advantage of contextual information provided by neighboring pixels along epipolar line \cite{yang2020cost}.

\subsection{Supervise on Cost Volume}
As shown in Fig.\ref{figure:multi-strategies}, it is hard to correct misestimated depth in refinement stages.
To attain better predictions before neighboring-based refinement, we further utilize the information in cost volume at $2^{nd}$ stage.

Cost volume is defined to reflect the similarity between different views, where the true depth value should have the lowest cost, which means the probability distribution should be unimodal and peaked at the true depth hypothesis under ideal circumstances.
Based on this assumption, we construct unimodal distributions as reference distributions which directly constraint on the cost volume to reduce errors introduced by multi-modal distributions.
Following \cite{zhang2020adaptive}, we defined reference unimodal distribution as:
\begin{equation}
	P^l(i)=softmax(-\dfrac{\vert d^l(i)-d^l_{gt}(i)\vert}{\sigma_{i}}),\label{eq3}
\end{equation}
where $\sigma_{i}$ is variance of reference distribution for pixel $i$, which controls the sharpness of peak and it is defined as:
\begin{equation}
	\sigma^l_i=\alpha_c^l(1-f^l_i)+\beta_c^l,
\end{equation}
where $f^l_i$ is confidence value of pixel $i$ in stage $l$.
We estimate confidence value for each pixel by a 2D confidence estimation network. $\alpha_c^l$ and $\beta_c^l$ are scale factor and lower bound, respectively.
Different from \cite{zhang2020adaptive}, we use learned neural network parameters instead of hand-crafted factors to adapt different properties of probability distributions for different datasets. 
Large $\sigma$ indicates low confidence of pixel, which usually caused by mismatch in textureless regions.

We leverage stereo focal loss proposed by AcfNet \cite{zhang2020adaptive} to guide network to generate unimodal distributions for each pixel.
The stereo focal loss is defined as:
\begin{equation}
	\mathcal{L}_{SF}=\dfrac{1}{\vert\mathcal{P}\vert}\sum_{i\in\mathcal{P} }(\sum_{d=0}^{D-1}(1-P_i(d))^{-\gamma}\cdot(-P_i(d)\cdot \mathrm{log}\hat{P}_i(d))),
\end{equation}
where $P_i(d)$ is probability value of reference unimodal distribution at depth $d$ of pixel $i$, and $\hat{P}_i(d)$ is estimated probability of pixel $i$ at depth $d$ given by our UNet.
Instead of simple cross entropy loss, we set $\gamma \ge 0$ to force unimodal guidance to focus on high-confidence regions.

After adaptive unimodal filtering (AUF), some local maximas are eliminated, and the errors in stage 2 are decreased.
Fig.\ref{figure:multi-strategies}(left) presents depth searching range of our method with and without AUF, respectively.
The depth sampling range in $2^{nd}$ stage indicates that AUF narrows down the sampling range and contributes to a more accurate initial depth map before refinement.

\subsection{Loss Function}
Our total loss consists of three parts: regression loss in each stage, stereo focal loss and confidence loss, which is denoted as:
\begin{equation}
	\begin{aligned}
		\mathcal{L}=&\lambda_{SF}\mathcal{L}_{SF}+\lambda_{C}\mathcal{L}_{C} +\sum_{l=1}^{L}\omega^l \mathcal{L}^l_{regression} \label{eq6}
	\end{aligned}
\end{equation}
Where $\lambda_{SF}$ and $\lambda_{C}$ are two factors to balance stereo focal loss and confidence loss on second stage.
The confidence loss $\mathcal{L}_{C}$ is defined as: 

\begin{equation}
	\begin{aligned}
		\mathcal{L}_{C}=\dfrac{1}{\vert\mathcal{P}\vert}\sum_{i\in\mathcal{P}}-\mathrm{log}f_i
	\end{aligned}
\end{equation}
We apply negative log-likelihood function as confidence loss to encourage confidence estimation network to predict high confidence values for each pixel.

Regression loss $\mathcal{L}^l_{regression}$ is defined to reflect the difference between the predicted depth map and ground-truth at stage $l$.
We use hand-crafted weight $\omega$ at each stage.
For stage $l$, the $L1$ norm is defined as:
\begin{equation}
	\begin{aligned}
		\mathcal{L}^l_{regression}=\sum_{i\in\mathcal{P}}\lVert d_i^l-\hat{d_i^l} \rVert_1
	\end{aligned}
\end{equation}

\section{Experiment}
\subsection{Dataset}
{\bf DTU Dataset.}
We train and evaluate our network on DTU dataset \cite{aanaes2016large} to obtain quantitative results.
DTU dataset \cite{aanaes2016large} consists of 124 large scale of scenes in 49 or 64 different views and 7 different light conditions, with the evaluation reference obtained by a structured light scanner.
We use the same splited training and evaluation sets with \cite{yao2020blendedmvs,yang2020cost}.
While the original size of evaluation image is $1600 \times 1200$, we crop it to $1600\times1184$ to fit the upsample process. 

{\bf BlendedMVS.}
BlendedMVS \cite{yao2020blendedmvs} is a collection of images captured from different views of 113 various scenarios.
It contains 17K training samples in low-resolution ($768\times576$) as well as high-resolution ($2048\times1536$).
Following the official training and validation list given by the released dataset files, we divided 106 scenes for training and the other 7 for validation in low-resolution BlendedMVS.
We train our model on low-resolution BlendedMVS and evaluate on both low-resolution and high-resolution.

\vspace{1mm}
\begin{minipage}{1\textwidth}
\begin{minipage}[t]{0.48\textwidth}
  \centering
     \makeatletter\def\@captype{table}\makeatother\caption{Quantitative results on DTU}
     \scriptsize{
        \begin{tabular}{|l|ccc|} \cline{1-4}
        \centering
            Methods & acc. &comp. &overall\\ \cline{1-4}
            MVSNet\cite{yao2018mvsnet} &0.396 &0.527 &0.462\\ 
    		R-MVSNet\cite{yao2019recurrent} &0.383 &0.452 &0.418\\
    		MVSCRF\cite{xue2019mvscrf} &0.371 &0.426 &0.398\\
    		PointMVSNet\cite{chen2019point} &0.361 &0.421 &0.391\\
    		CVP-MVSNet\cite{yang2020cost} &{\bf0.296} &0.406 &0.351\\
    		AACVP-MVSNet\cite{yu2021attention} &0.357 &0.326 &0.341\\
    		Vis-MVSNet\cite{zhang2020visibility} &0.369 &0.361 &0.365\\
    		USCNet\cite{mao2021uasnet} &0.338 &0.349 &0.344\\
    		PVSNet\cite{xu2020pvsnet} &0.337 &0.315 &{\bf0.326}\\
    		Ours &0.379 &{\bf0.278} &0.328\\ \cline{1-4}
        \end{tabular} \label{tab:dtu}}
\end{minipage}
\begin{minipage}[t]{0.45\textwidth}
        \makeatletter\def\@captype{table}\makeatother\caption{Different strategies}
        \scriptsize{
         \begin{tabular*}{\linewidth}{|l|ccc|}\cline{1-4} 
		\centering
			Strategies &acc. &comp. &overall\\ \cline{1-4}
			DHS1 &0.444 &0.361 &0.402 \\
			DHS1+DHS2 &{\bf 0.338} &0.349 &0.344 \\
			DHS1+DHS3 &0.404 &0.321 &0.362 \\
			DHS1+DHS2+DHS3 & 0.389 &{\bf 0.279} & {\bf 0.334} \\
			\cline{1-4}  
		\end{tabular*} \label{tab:strategies}
		\\
		\parbox{5.2cm}{Note: DHS1 denotes uniformly sampling the whole range, DHS2 denotes variance-based method, and DHS3 back-projects pixels along epipolar line to calculate depth searching range.
		 }
		}
   \end{minipage}
\end{minipage}
\vspace{1mm}

\subsection{Implementation details}
{\bf Training.}
We train and evaluate our model on DTU dataset and low-resolution BlendedMVS.
For first stage, we uniformly sample the whole depth range $[425,1065]$ with $D^1=48$, while for 2nd and 3rd stage, we choose $D^2=32$ and $D^3=8$, respectively.
As the training process with high-resolution inputs is memory and time consuming, we downsample the training set into a size of $320\times256$, and the coarsest resolution is $40\times32$ in the first stage.
We set hyperparameters $\lambda_{SF}=10$, $\lambda_{C}=80$ in equation \eqref{eq6} and choose $\omega^1=0.5$, $\omega^2=1$, $\omega^3=2$ to balance $L1$ loss in each stage.
As for the reference unimodal distribution, the scale factors are initialized as $\alpha_c^2=13$ and $\beta_c^2=9$, respectively, based on empirical evidence from \cite{zhang2020adaptive}.
We use 3 different views as inputs and Adam \cite{kingma2014adam} as optimizer in the training stage of the proposed network.
We set batch size as 16 and train our model on 2 Nvidia GeForce RTX 3090 for 40 epoches with initial learning rate 0.001 multiplied by 0.5 at 10th, 12th, 14th, 20th epoch. \par
{\bf Evaluation.}
For DTU dataset,we crop the original images to $1600\times1184$ for evaluation.
We set $L=5$ for image feature pyramid to maintain a similar size with training stage at the coarsest stage ($50\times37$).
Similar to \cite{yao2018mvsnet,yao2019recurrent,yang2020cost}, we choose 5 views in evaluation for fair comparison.
The depth sampling numbers $D$ in each stage the same as training process.
As for BlendedMVS, we evaluate our proposed method on both low-resolution and high-resolution dataset. \par
{\bf Post processing and Metrics.}
After estimating the depth map, we fuse all views into a dense point cloud model for each scene.
For fair comparison, we follow the common post processing method used by \cite{yao2018mvsnet,yao2019recurrent,yang2020cost}, which is a fusion method provided by Galliani et al. \cite{galliani2015massively}.
We run the official evaluation code provided by DTU dataset \cite{aanaes2016large} to obtain quantitative results of mean accuracy (acc.), mean completeness (com.) and overall score (overall).
The evaluation results are listed in Tab.\ref{tab:dtu}.
\begin{figure}
	\begin{minipage}[t]{0.52\linewidth}
		\centering
		\includegraphics[width=3.5in]{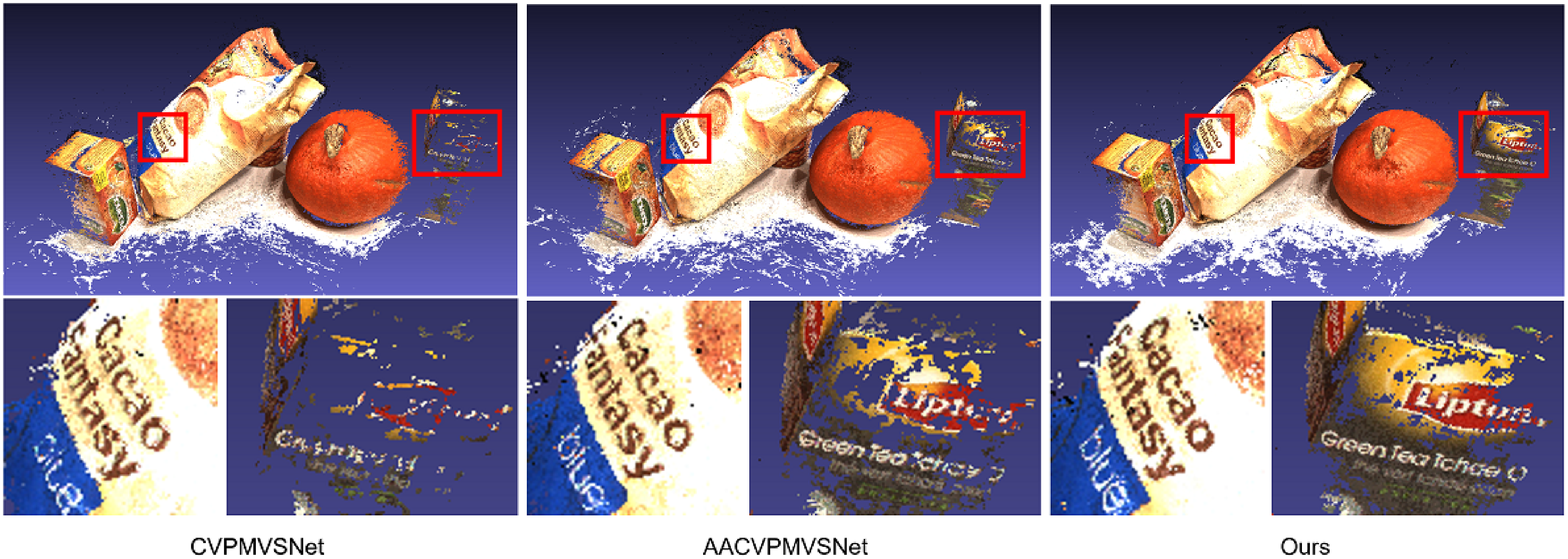}
	\end{minipage}
	\begin{minipage}[t]{0.44\linewidth}
		\centering
		\includegraphics[width=3.3in]{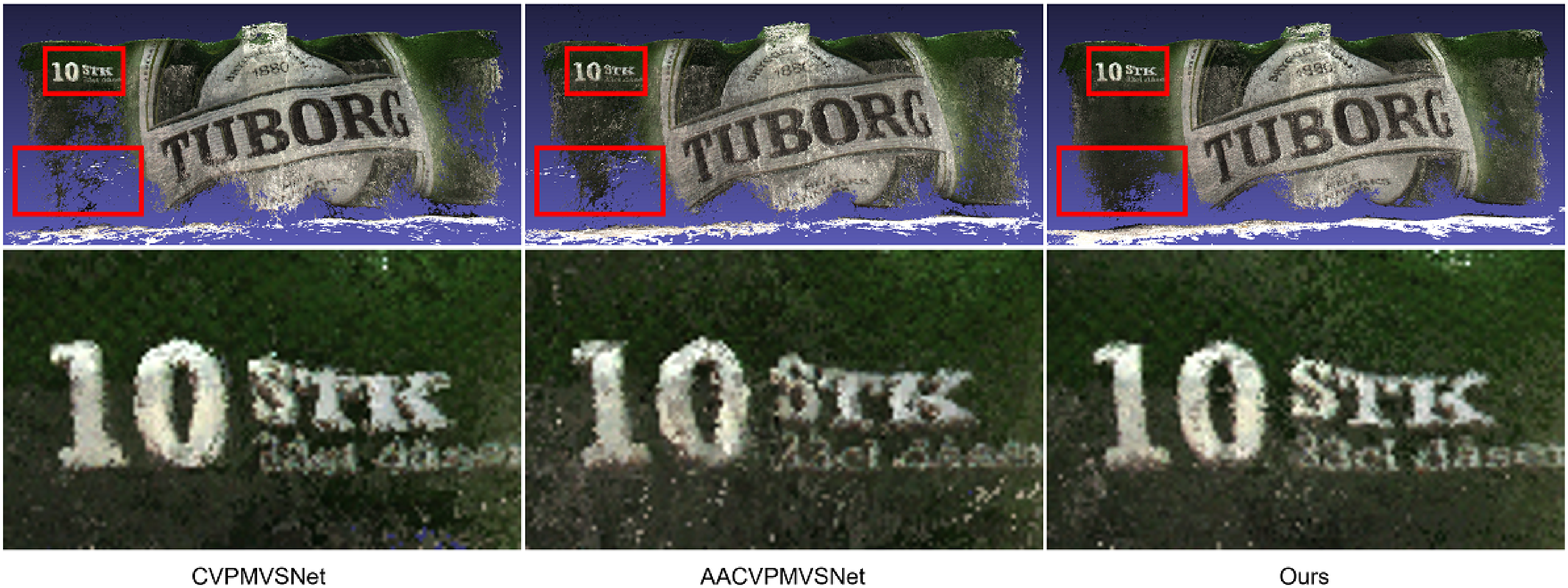}
	\end{minipage}
	\caption{Reconstruction results on DTU dataset.}
	\label{figure:dtu}
\end{figure}

\subsection{Results on DTU dataset}
We train and evaluate our method on DTU dataset to conduct quantitative results in comparison with other learning based methods.
As shown in Tab.\ref{tab:dtu}, our method achieves state-of-the-art results in overall score, which is comparable to PVSNet \cite{xu2020pvsnet}.
Especially, our method outperforms all methods in Tab. \ref{tab:dtu} in terms of completeness.
As shown in Fig. \ref{figure:dtu}, We visualize several reconstructed 3D models constructed by CVP-MVSNet \cite{yang2020cost}, AACVP-MVSNet \cite{yu2021attention} and our proposed method.

\subsection{Results on BlendedMVS}
\begin{figure}
	\begin{minipage}[t]{0.48\linewidth}
		\centering
		\includegraphics[width=2.8in]{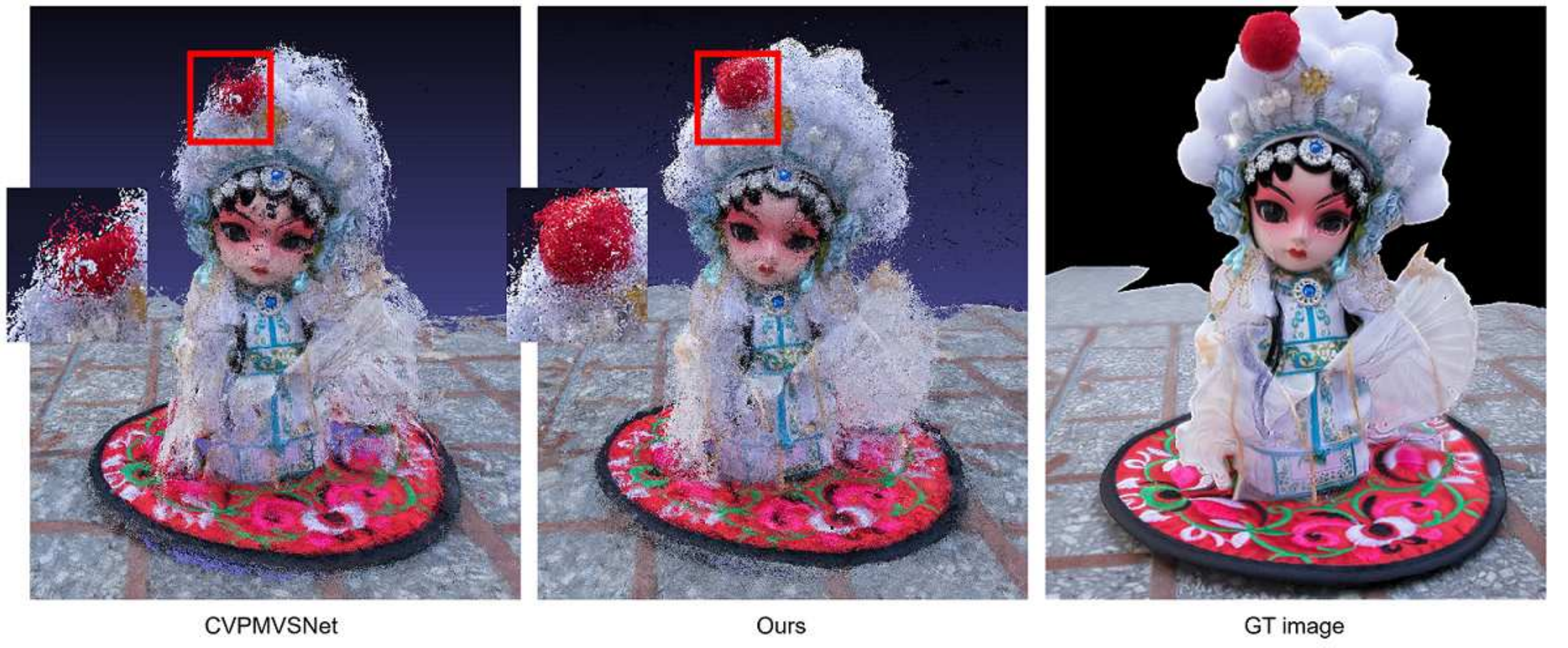}
	\end{minipage}
	\begin{minipage}[t]{0.48\linewidth}
		\centering
		\includegraphics[width=2.8in]{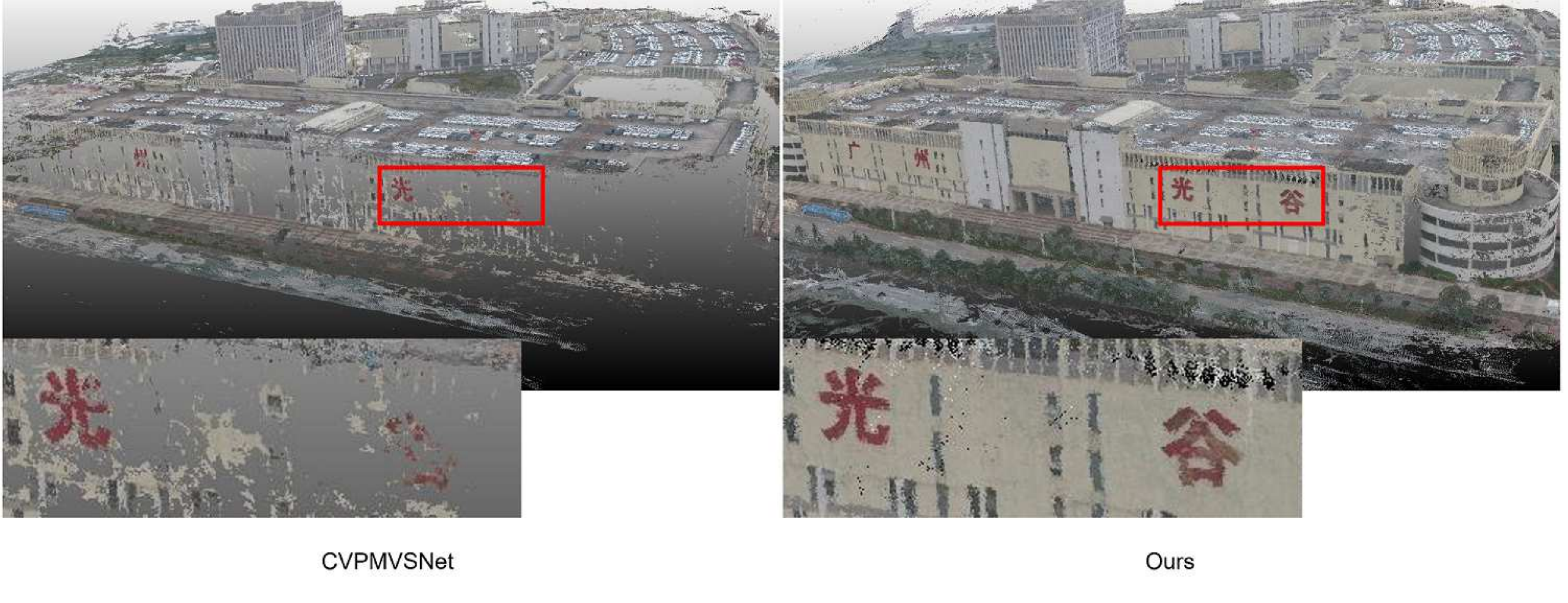}
	\end{minipage}
	\caption{Reconstruction results on BlendedMVS dataset.}
	\label{figure:blended}
\end{figure}

As BlendedMVS dataset does not provide any reference point clouds for quantitative evaluation, we conduct the visual comparison with CVP-MVSNet \cite{yang2020cost}.
$L=3$ in training process, while for evaluation, we set $L=5$ and $L=6$ for low and high resolution evaluation sets, respectively.
In the same way, we compare our method with CVP-MVSNet \cite{yang2020cost} and the results of low- and high-resoluion dataset are shown in Fig.\ref{figure:blended}.
On high-resolution data sets, the superiority of our method in terms of completeness is even more evident.

\subsection{Effectiveness of Multi-strategies}
As shown in Tab.\ref{tab:strategies}, we compare our proposed multi-strategies with other combinations of strategies. 
For strategy "DHS1", we apply DHS1 and uniformly sample at each stage with handcrafted searching range $[40,20,10,5]$ from $2^{nd}$ to $5^{th}$ stage (range 1 corresponds to stage 2).
"DHS1+DHS2" and "DHS1+DHS3" perform results of single strategy DHS2 and DHS3 from $2^{nd}$ to $5^{th}$ stage, respectively.

As shown in Fig.\ref{figure:multi-strategies}, CVPMVSNet \cite{yang2020cost} applies DHS3 in each stage and fails to locate an interval which contains true depth value from $2^{nd}$ to the last stage.
Its single and inflexible range searching strategy makes it hard to jump out of the pattern and rectify mismatch in previous stage.
We believe that DHS2 which is based on the variance of previous prediction is more accurate and effective to locate true depth value (see (b) range 1 in Fig.\ref{figure:multi-strategies}), but proper scale factors are needed in each specific stage.
Our proposed multi-strategies method combines both DHS2 and DHS3, in the second stage, DHS2 gives a reasonable searching range based on previous predicted probabilities, while for the rest stages, DHS3 which is parameter-free provides an effective way to propagate depth of neighboring pixels along epipolar line to an arbitrary resolution during upsampling refinement.
Noteworthy, multi-strategies method works better when it combines with two separated UNets (see CVP-MS and CVP-MS-$U^2$Net in Tab.\ref{tab:strategies}).

\tabcolsep 4pt
\begin{table*}[!htb]
\centering
	\caption{\label{tab:ablation} Ablation study on DTU dataset}
	{\scriptsize{
		\begin{tabular}{|l|cccc|ccc|}\cline{1-8} 
			Methods &Variance  &Epipolar line  & $U^2$Net &Auf & acc. &comp. &overall\\
			\cline{1-8}
			CVP(baseline) &\XSolid &\Checkmark &\XSolid &\XSolid &{\bf 0.313} &0.394 &0.354\\
			CVP-MS &\Checkmark &\Checkmark &\XSolid &\XSolid &0.343 &0.439 &0.391\\
			CVP-$U^2$Net &\XSolid &\Checkmark &\Checkmark &\XSolid&0.330 &0.379 &0.355\\
			CVP-MS-Auf &\Checkmark &\Checkmark &\XSolid &\Checkmark&0.321 &0.398 &0.360\\
			CVP-MS-$U^2$Net &\Checkmark &\Checkmark &\Checkmark &\XSolid &0.389 &0.279 &0.334\\
			Ours &\Checkmark &\Checkmark &\Checkmark &\Checkmark &0.379 &{\bf0.278} &{\bf0.328}\\
			\cline{1-8}  
		\end{tabular}   }
		\\
		\vspace{1mm}
	} 
\end{table*}
\vspace{-2mm}

\subsection{Ablation study}
In this section, we perform ablation experiments on DTU dataset to validate the effectiveness of each component of our proposed network.
Results are shown in Tab.\ref{tab:ablation}.
Below we analyse each component in details.

$\bullet$ {\bf Non-parameter-sharing UNet.}
3D UNet is designed for cost volume regularisation and explore cost volume information in three dimensions.
Quantitative results on DTU dataset show that our two parameter-separating UNets gain better results (0.328 vs.0.360) than parameter-sharing UNet.
The huge gap indicates that former stages which search in a wider range have different characteristics with refinement stages in the cost volume regularization process.\par

$\bullet$ {\bf Supervise on cost volume.}
While multi-strategies with two non-parameter-sharing UNet framework has achieve promising results (see CVP-MS-$U^2$Net in Tab.\ref{tab:ablation}), we obtain even better results when further adding adaptive unimodal filtering (AUF) on $2^{nd}$ stage.
As shown in Fig.\ref{figure:multi-strategies}, the depth sampling range in $2^{nd}$ stage is narrowed after adding AUF module.
Interestingly, quantitative results of CVP-MS-Auf and CVP-MS in Tab.\ref{tab:ablation} show that adaptive unimodal filtering gives a greater boost when parameter-sharing UNet is adopted.
\vspace{1mm}
\tabcolsep 3pt
\renewcommand\arraystretch{1.3}
\begin{table*}[!htb]
	\centering
	\caption{\label{tab:res} Quantitative results on DTU dataset with different training and evaluation resolution.}
	{\scriptsize{
	\centering
		\begin{tabular}{|ccc|cc|ccc|}\cline{1-8} 
			Coarsest $Res_{T}$ &Coarsest $Res_{E}$ &$Levels_{E}$ &mem.(M) &runtime(s) &acc. &comp. &overall  \\ \cline{1-8} 
			$40\times32$ &$25\times18$ &6 &\multirow{2}*{{\bf6809}} & \multirow{2}*{2.543} &0.372 &0.292 &0.332\\ 
			$20\times16$ &$25\times18$ &6 & & &0.382 &0.324 &0.353 \\ \cline{1-8}
			$40\times32$ &$50\times37$ &5 &\multirow{2}*{7863} & \multirow{2}*{2.550} &0.379 &{\bf0.278} &{\bf0.328}\\ 
			$20\times16$ &$50\times37$ &5 & & &0.371 &0.328 &0.349 \\ \cline{1-8}
			$40\times32$ &$100\times74$ &4 &\multirow{2}*{6935} & \multirow{2}*{2.483} &0.360 &0.311 &0.335\\
			$20\times16$ &$100\times74$ &4 & & &{\bf0.349} &0.478 &0.413 \\  \cline{1-8}
			$40\times32$ &$200\times148$ &3 &\multirow{2}*{7861} & \multirow{2}*{{\bf2.366}} &0.375 &0.530  &0.452\\
			$20\times16$ &$200\times148$ &3 & & &0.531 &1.959 &1.245 \\
			\cline{1-8} 
		\end{tabular}
		}
	}
\end{table*}

$\bullet$ {\bf Image resolution during training and evaluation.}
Tab. \ref{tab:res} shows that the performance of the model trained with higher resolution input is better than that with lower resolution input.
To discover the relationship between pyramid levels and quality of output depth map, we also evaluate our method with different pyramid levels on DTU dataset.
As shown in Tab. \ref{tab:res}, coarse-to-fine network with 5 pyramid stages achieves the best overall score.

\section{Conclusion}
In this paper, we present an efficient deep-learning based cost volume pyramid network for MVS.
By combining different sampling range estimation strategies for each stage, we integrate multi-dimensional information without additional neural network modules.
Then, we apply adaptive unimodal filters to further improve the low-resolution depth map before refinement.
Results on different datasets show the effectiveness and generalisability of our method.


\bibliographystyle{unsrtnat}
\bibliography{references}  






\end{document}